\documentclass[letterpaper, 10 pt, conference]{ieeeconf}

\IEEEoverridecommandlockouts

\usepackage{graphicx}
\usepackage{multirow}
\usepackage{multicol}
\usepackage{url}
\usepackage{amsmath}
\usepackage{amsfonts}
\usepackage{subcaption}

\usepackage{tikz}
\usepackage{lipsum}
\usepackage{hyperref}
\newcommand\preprinttext{%
    \scriptsize \textbf{[This is a preprint - Published in IEEE/RAS International Conference on Automation Science and Engineering 2025 - DOI: \href{https://doi.org/10.1109/CASE58245.2025.11163906}{10.1109/CASE58245.2025.11163906}]}
}
\newcommand\copyrighttext{
    \scriptsize \textcopyright 2025 IEEE. Personal use of this material is permitted. Permission from IEEE must be obtained for all other uses, in any current or future media, including reprinting/republishing this material for advertising or promotional purposes, creating new collective works, for resale or redistribution to servers or lists, or reuse of any copyrighted component of this work in other works.
}
\newcommand\noticeblock{
    \begin{tikzpicture}[remember picture,overlay]
    \node[anchor=north,yshift=-30pt] at (current page.north) {\preprinttext};
    \node[anchor=south,yshift=15pt] at (current page.south) {
        \parbox{\dimexpr\textwidth-\fboxsep-\fboxrule\relax}{\copyrighttext}
    };
    \end{tikzpicture}
}

\title{\LARGE \bf
    Multi-stage Planning for Multi-target Surveillance using Aircrafts Equipped with Synthetic Aperture Radars Aware of Target Visibility
}

\author{
    Daniel Fuertes$^{1}$,
    Carlos R. del-Blanco$^{1}$,
    Fernando Jaureguizar$^{1}$,
    Juan José Navarro-Corcuera$^{2}$,
    Narciso García$^{1}$
    \thanks{
        $^{1}$Daniel Fuertes, Carlos R. del-Blanco, Fernando Jaureguizar, and Narciso García are with Grupo de Tratamiento de Imágenes, Information Processing and Telecommunications Center, ETSI Telecomunicación, Universidad Politécnica de Madrid, 28040 Madrid, Spain.
        {\tt\small \{d.fcoiras, carlosrob.delblanco, fernando.jaureguizar, narciso.garcia\}@upm.es}
    }
    \thanks{
        $^{2}$Juan José Navarro-Corcuera is with Airbus Defence and Space, 28906 Madrid, Spain.
        {\tt\small juan.j.navarro@airbus.com}
    }
}

\begin{document}

% Style
\maketitle
\thispagestyle{empty}
\pagestyle{empty}
\noticeblock

% Abstract
\begin{abstract}
Generating trajectories for synthetic aperture radar (SAR)-equipped aircraft poses significant challenges due to terrain constraints, and the need for straight-flight segments to ensure high-quality imaging. Related works usually focus on trajectory optimization for predefined straight-flight segments that do not adapt to the target visibility, which depends on the 3D terrain and aircraft orientation. In addition, this assumption does not scale well for the multi-target problem, where multiple straight-flight segments that maximize target visibility must be defined for real-time operations. For this purpose, this paper presents a multi-stage planning system. First, the waypoint sequencing to visit all the targets is estimated. Second, straight-flight segments maximizing target visibility according to the 3D terrain are predicted using a novel neural network trained with deep reinforcement learning. Finally, the segments are connected to create a trajectory via optimization that imposes 3D Dubins curves. Evaluations demonstrate the robustness of the system for SAR missions since it ensures high-quality multi-target SAR image acquisition aware of 3D terrain and target visibility, and real-time performance. % due to fast computation speed.
\end{abstract}

% Sections
\section{Introduction}
\label{sec:intro}

% Intro
Autonomous path planning for aircrafts equipped with a synthetic aperture radar (SAR) is a crucial challenge in various applications, including search-and-rescue, disaster management, military and defense operations, environmental monitoring, navigation and mapping, agriculture, oceanography, geology, mining, etc. The ability of SAR to capture high-resolution images under diverse weather conditions and during both day and night makes it particularly suitable for missions requiring robust and reliable target detection. However, SAR imaging quality is highly dependent on the trajectory of the aircraft, as it must follow a well-defined, straight-line flight path (segment) at an appropriate altitude to ensure optimal image resolution and target detection \cite{Cruz2022}. An additional challenge arises from the need to maintain proper target visibility, as terrain elevations can obstruct the line-of-sight to the target. Visibility is also influenced by aircraft altitude: while higher altitudes may degrade image quality and reduce detection accuracy, lower altitudes can introduce visibility constraints due to terrain orography and obstacles. Furthermore, when multiple targets must be observed, the aircraft must efficiently connect the predicted straight-flight segments (hereafter referred to simply as "segments" for simplicity) to minimize travel time and energy consumption while maintaining mission effectiveness.

% Solution
\begin{figure}[t] % thpb
  \centering
  \includegraphics[width=\linewidth]{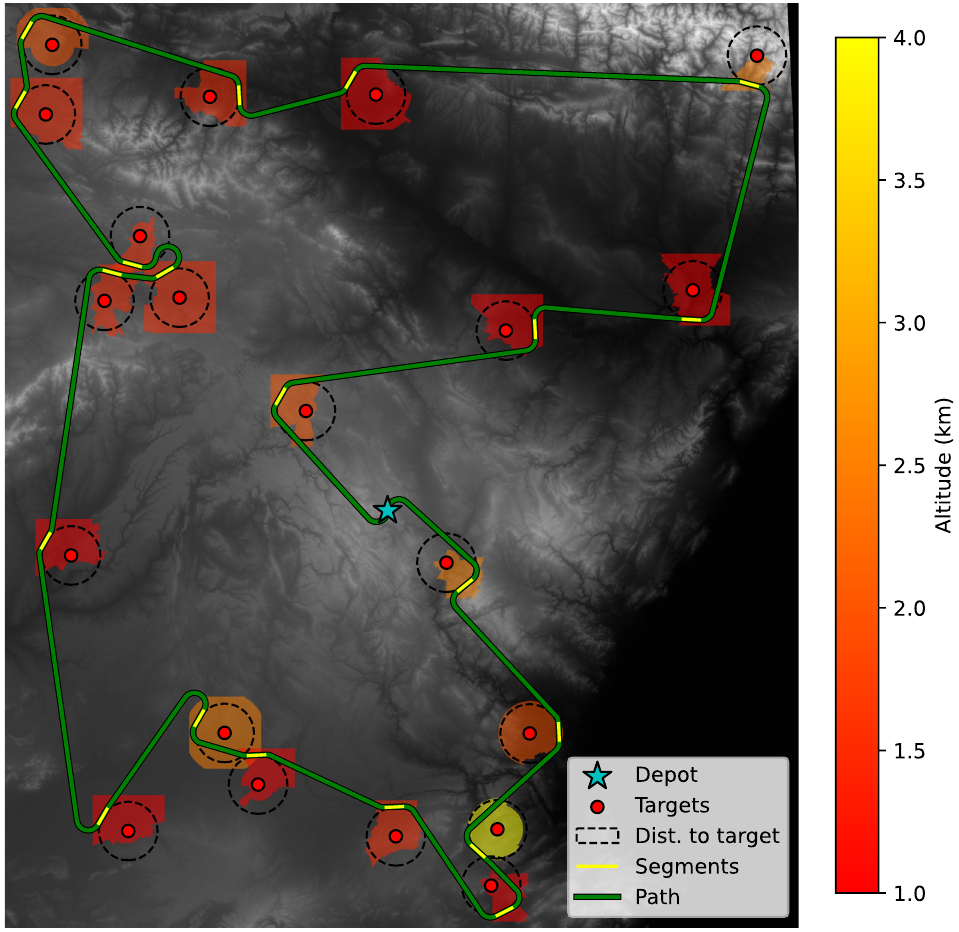}
  \caption{Estimated solution in a muti-target scenario. Dashed lines show the minimum distance to targets (red circles). The path (green lines) starts and ends at the depot (cyan star), connecting all straight-flight segments (yellow lines). Color-coded maps indicate target visibility at a specific altitude.}
  \label{fig:solution}
\end{figure}

% Workflow
\begin{figure*}[t] % thpb
  \centering
  \includegraphics[width=\textwidth]{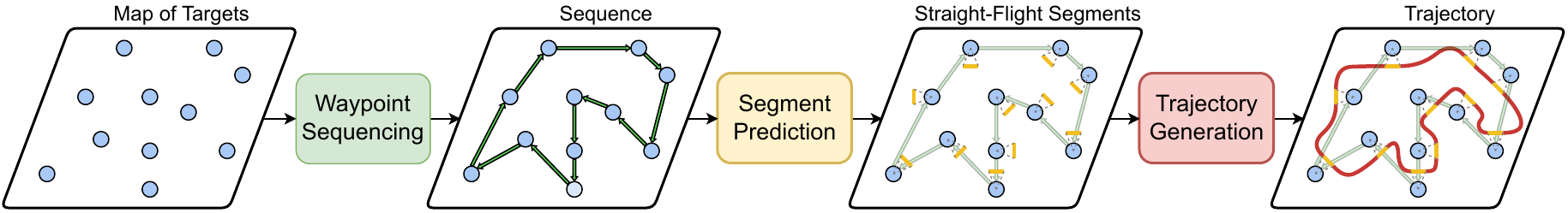}
  \caption{Workflow of the proposed system. The input is a map of targets (blue circles), which are ordered (green lines) using the waypoint sequencing module. After that, the straight-flight segments (yellow lines) are predicted considering the terrain elevations. Finally, the trajectory of the aircraft (red lines) is inferred from the waypoint sequence and the segments.}
  \label{fig:workflow}
\end{figure*}

% State of the art
Determining the optimal flight segment (characterized by its orientation, altitude, and duration) that maximizes the visibility for each potential target while simultaneously estimating the most efficient trajectory that connects multiple targets and minimizes resource consumption is a very complex challenge. The state of the art in this field is extensive and highly diverse. Many studies, such as \cite{Lahmeri2022, Luna2022}, focus on the coverage path planning problem, aiming to maximize the area captured by the SAR rather than addressing the multi-target optimization problem. Other approaches \cite{Sun2024, Lu2022} simplify the problem by ignoring terrain elevation constraints, focusing in the trajectory optimization. Works such as \cite{Sun2023, Xu2021} account for terrain elevations but restrict the problem to estimate the best trajectory for a single target. In addition, few studies tackle the multi-target problem specifically for SAR-equipped aircrafts. For instance, \cite{Li2018} extracts segments for SAR imaging imposing rectangular regions of interest. Then, the segments are connected using an A*-based search algorithm. Alternatively, \cite{Stecz2020} formulates the multi-target problem as a vehicle routing problem, solving it through linear optimization and refining the trajectories with Dubins curves.

% Disadvantages of state of the art
A key limitation of the aforementioned approaches is their reliance on predefined segment positions rather than actively predicting the most suitable ones that maximize target visibility. Instead, they mainly focus on providing efficient trajectories that connect the segments. However, determining automatically the segments that maximize target visibility in the acquired SAR imagery is crucial. Moreover, this capability becomes essential in scenarios involving dozens or even hundreds of targets, where manual selection is not feasible. Although few studies, such as \cite{Sanchez2022}, introduce visibility-based heuristics for efficient route planning, they focus on standard optical cameras and do not address the unique constraints of SAR imaging, which impose straight-line flight segments for proper SAR image acquisition.

% Paper proposal
To address these limitations, this paper presents a multi-stage path planning system for multi-target surveillance specifically designed for SAR-equipped aircraft that maximizes the visibility of targets in the SAR imagery. It integrates sequencing of multiple waypoints (targets), flight segment prediction with visibility awareness, and trajectory generation into a unified framework. For this purpose, a novel neural network for segment prediction that considers terrain elevation data to maximize target visibility at the lowest possible altitude is proposed. This approach eliminates the need for predefined segment positions and enhances adaptability to highly complex environments. The system is evaluated with a benchmark that assesses and compares the performance of multiple configurations of the SAR path planning system.

% Workflow
The system is divided into three key stages. The first stage is the waypoint sequencing, which determines the optimal visiting order of targets using a Transformer-based neural network \cite{Kool2019} that process a graph representation of the problem, ensuring efficient high-level route planning. The second stage is the segment prediction, built upon a MobileNetv3 \cite{Howard2019} backbone, which analyzes terrain elevations to select the best segments for SAR imaging. Finally, the last stage is the trajectory generation module that employs A* search \cite{Saadi2022} with 3D Dubins curves to compute feasible paths that connect the selected segments while maintaining a smooth aircraft motion. The proposed system offers a holistic and effective solution to the SAR image acquisition problem in multi-target scenarios. Fig. \ref{fig:solution} shows an example of performance of the system.

\section{Methodology}
\label{sec:system}

% Network
\begin{figure*}[t] % thpb
  \centering
  \includegraphics[width=0.8\linewidth]{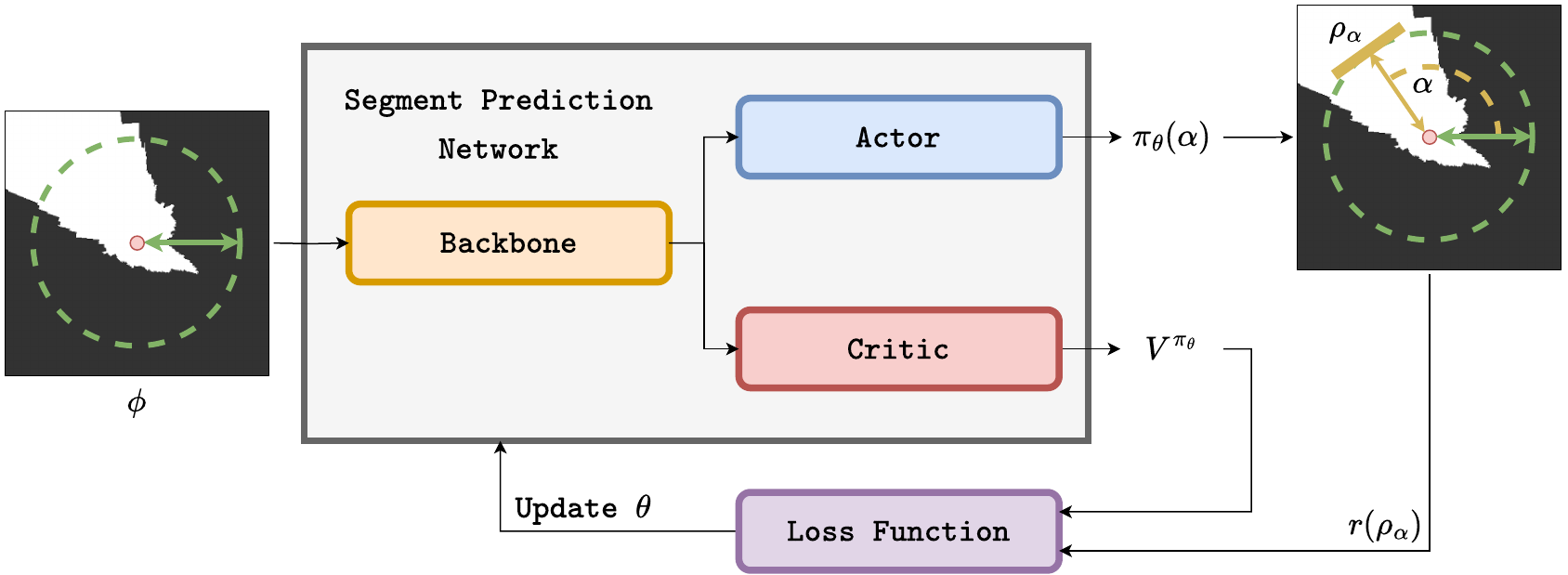}
  \caption{Segment prediction on a visibility map. Occluded areas are shown in black and visible ones in white. The map is processed by a feature extractor backbone, which feeds the actor module that predicts the policy $\pi_\theta(\alpha)$ used to infer segment $\rho_\alpha$. The critic module outputs the critic value $V^{\pi_\theta}$ to help training converge. During training, both $V^{\pi_\theta}$ and the reward $r(\rho_\alpha)$ (calculated from the segment visibility $\phi(\rho_\alpha)$) are used to update the model weights $\theta$ through the loss function.}
  \label{fig:network}
\end{figure*}

% Intro
The workflow of the proposed multi-stage system is illustrated in Fig. \ref{fig:workflow}. Notice how the final generation of trajectories depends on the estimations of both the waypoint sequence and the segment predictions. The following subsections describe in detail each of these three stages.

% Waypoint Sequencing
\subsection{Waypoint Sequencing}
\label{sec:waypoint}

% Intro
The waypoint sequencing problem is formulated as a Traveling Salesman Problem (TSP) since both share the objective of determining the optimal sequence of waypoints that minimizes the total route cost. Given a set of $n$ targets $\mathcal{W} = \{w_1, \ldots, w_n\}$, the aim is to find a permutation $\sigma$ of waypoints that minimizes the total tour length, expressed as

% Objective
\begin{equation}
    \min_{\sigma} \sum_{i=1}^{n-1} ||w_{\sigma(i)} - w_{\sigma(i+1)}||_2,
\end{equation}

% Objective
where $w_{\sigma(i)}$ represents the $i$-th waypoint in the sequence defined by $\sigma$, and $||w_i - w_j||_2$ denotes the travel cost between two waypoints.

% Transformer
To efficiently solve this problem, we employ an autoregressive Transformer-based neural network inspired by \cite{Kool2019}, which has the advantage of having a significant reduction in computational cost compared to standard approaches based on mathematical optimization and linear programming. The model processes the whole set of waypoints in parallel, leveraging the self-attention mechanism \cite{Vaswani2017}, to encode the relationships of the underlying graph defined by the set of targets. Training is performed using deep reinforcement learning, where the Transformer model is optimized under an Actor-Critic framework. The actor network, parameterized by the set of weights $\varphi$, represents the waypoint selection policy $\pi_\varphi$, while the critic network estimates the expected cost value $V^{\pi_\varphi}$ of a tour. The ultimate objective is to minimize the expected tour cost given by

% Loss function
\begin{equation}\label{eq:loss}
    \mathcal{L} = - \mathbb{E} \left[ A \log \pi_\varphi(\sigma) \right] + \mathcal{L}_{\text{critic}} - \lambda H
\end{equation}

% Loss function
where $A = \sum_{i=1}^{n} ||w_{\sigma(i)} - w_{\sigma(i+1)}||_2 - V^{\pi_\varphi}$ is the advantage function, $H$ is the entropy regularization term weighted by the value $\lambda=0.01$ as in \cite{Neumann2023}, and $\mathcal{L}_{\text{critic}}$ is the critic loss, computed as the Mean Squared Error (MSE) between the predicted $V^{\pi_\varphi}$ and actual $r(\sigma)$ tour costs:

% Critic loss
\begin{equation}
    \mathcal{L}_{\text{critic}} = \text{MSE}(V^{\pi_\varphi}, r(\sigma))
\end{equation}

% Critic loss
As a result of the Transformer-based Actor-Critic framework that improves the training converge, an efficient waypoint sequencing prediction is obtained.

% Segment prediction
\subsection{Segment Prediction}
\label{sec:segment}

% Visibility maps
Segment prediction is guided by the visibility of the target at a given aircraft altitude. For each target, multiple visibility maps, denoted by $\phi$, are extracted in parallel from the terrain information at altitudes ranging from $1~$km to $4~$km in steps of 500 meters. Each visibility map is computed by identifying the terrain regions with a clear line of sight to the target. A square region from the terrain model centered on the target is extracted, whose elevation data are adjusted for Earth's curvature. A point $p$ in the squared region is considered visible if no elevation of the terrain obstructs the view (clear line of sight), assigning a reward $r(p)$ equal to $\phi(p)=1$ and $\phi(p)=0$ otherwise, since the visibility map is binary.

% Segments
Then, every visibility map is fed into a segment prediction network (see Fig. \ref{fig:network}) that uses a MobileNetv3 \cite{Howard2019} architecture as backbone. This generates feature embeddings that are processed by actor and critic subnetworks, implemented by two linear projection layers that predict the policy distribution $\pi_\theta$ and the critic value $V^{\pi_\theta}$, respectively, where $\theta$ denotes all the weights of the segment prediction network. The actions, represented by $\alpha$, are sampled from $\pi_\theta$ and correspond to the predicted angles. Notably, the angle $\alpha$ defines the orientation of a radius that is perpendicular to the desired segment, denoted by $\rho_\alpha$, which is tangent to a circumference centered at the target. This circumference has a fixed radius of 1.5 km for the experiments (see Section \ref{sec:results}), although other values could be tested, and represents the minimum distance required for the aircraft to approach the target during SAR detection. The length of the segment depends on both the observation time and the aircraft's speed, and its endpoints are computed from $\alpha$ via basic trigonometric transformations.

% Rewards
Once the segment $\rho_\alpha$ is determined, the overall reward for the segment is calculated as the sum of the rewards (i.e. their visibility) at all points along the segment:

% Rewards
\begin{equation}
    r(\rho_\alpha) = \sum_{\forall p \in \rho_\alpha} \phi(p)
\end{equation}

% Critic loss
Notice that multiple segments are predicted for all the altitudes considered. Among all these segments, the one corresponding to the lowest altitude that exhibits the highest overall reward/visibility is selected (notice that visibility cannot decrease as the altitude grows), as it provides optimal conditions for SAR detection. The critic loss is then defined similar to that in Section \ref{sec:waypoint}:

% Critic loss
\begin{equation}
    \mathcal{L}_{\text{critic}} = \text{MSE}(
        V^{\pi_\theta}, \,
        r(\rho_\alpha)
    )
\end{equation}

% Advantage function
As in Eq. \ref{eq:loss}, the Actor-Critic loss incorporates the advantage function, defined in this case as $A = r(\rho_\alpha) - V^{\pi_\theta}$.

% Trajectory genetation
\subsection{Trajectory Generation}
\label{sec:trajectory}

% Intro
The trajectory generation module computes an optimized flight path by integrating the waypoint sequence (obtained from the first stage, Transformer-based TSP solver) and the segments (yielded by the segment prediction network, second stage). This path planning problem is modeled as a graph search, where feasible paths between the end point of a segment and the starting point of another (related to different waypoints) are determined using A* with 3D Dubins curves to ensure smooth and feasible trajectories.

% Cost
Given the set of $n$ waypoints $\mathcal{W}$, the objective is to calculate a trajectory that minimizes the total cost of the path, considering the motion constraints of the aircraft, and visiting every predicted segment (associated with a target). The A* algorithm estimates which point of the two that define a segment is the starting point (the ending one is just the other) to minimize the total cost of the path. The cost function $g(\tau)$ evaluates a candidate trajectory $\tau$ as:

% Cost
\begin{equation}
    g(\tau) = \sum_{i=1}^{n-1} d_\tau \left( b_{\sigma(i)}, a_{\sigma(i+1)} \right),
\end{equation}

% Cost
where $b_{\sigma(i)}$ is the exit point of each previous segment, $a_{\sigma(i+1)}$ is the chosen entry point for each next segment, and $d_\tau \left( b_{\sigma(i)}, a_{\sigma(i+1)} \right)$ indicates the length of the 3D Dubins curve connecting $b_{\sigma(i)}$ and $a_{\sigma(i+1)}$. The cost reflects the transition distance between segments, ensuring that A* selects paths that minimize the overall travel length.

% Heuristic
To speed up the path search, the following heuristic function $h(s)$ is used in A*, which efficiently estimates the remaining path cost from the current state $s$ to the goal:

% Heuristic
\begin{equation}
    h(k) = \sum_{i=k}^{n-1} ||w_{\sigma(i)} - w_{\sigma(i+1)}||_2,
\end{equation}

% Heuristic
where $k$ is the current waypoint index. This heuristic, based on the straight-line distance between waypoints, is admissible and ensures optimality in the absence of obstacles. Notice that it estimates the Euclidean distances to the following targets to visit and not to their segments, which ensures underestimating the optimal cost value.

% Total
Finally, the search proceeds iteratively, expanding the most promising nodes based on the A* cost function:

% Total
\begin{equation}
    f(k) = g(k) + h(k),
\end{equation}

% Total
where $g(k)$ accumulates the cost from the start, and $h(k)$ estimates the remaining cost. This formulation allows for efficient generation of feasible paths, while optimizing waypoint sequencing and respecting aircraft motion constraints.

\section{Results}
\label{sec:results}

% Dataset
The proposed path planning method for SAR missions was evaluated using a dataset generated from 54,000 randomly sampled targets on a terrain elevation map, resulting in 54,000 visibility maps. The dataset was divided into 50,000 samples for training, 2,000 for validation, and 2,000 for testing. For the evaluation of the system, scenarios with $n \in \{20, 50, 100\}$ targets were generated by randomly sampling the targets from the test set. This follows the experimental setup of other works \cite{Kool2019,Fuertes2023}. In addition, a depot location was placed in the center, as illustrated in Fig. \ref{fig:solution}.

% Experiments
For the segment prediction network, several backbone architectures have been compared: MobileNetv3 \cite{Howard2019}, ConvNext \cite{Liu2022}, EfficientNetv2 \cite{Tan2021}, RegNet \cite{Radosavovic2020}, MaxViT \cite{Tu2022}, SwinT \cite{Liu2022b}, and ViT \cite{Dosovitskiy2021}. The proposed networks have been trained using the Adam optimizer with an initial learning rate of $10^{-4}$, a batch size of 6, and visibility maps of size $300 \times 300$, incorporating data from seven different altitudes for each target. The evaluation metrics is the segment accuracy that quantifies the proportion of segment points that maintain a clear line of sight to the target:

% Accuracy
\begin{equation}
    \text{Accuracy} = \frac{
        \sum_{p \in \rho_\alpha} \phi(p).
    }{
        |\rho_\alpha|
    }
\end{equation}

Regarding the waypoint sequencing, three categories of solvers have been evaluated: linear optimizers such as OR-Tools \cite{Ortools2024}, Gurobi \cite{Gurobi2024} (limited to 60 seconds for large $n\geq50$ to ensure solutions in tractable time), Concorde \cite{Concorde2006}, and Truncated Branch-and-Bound (B\&B) \cite{Zhang2023}; heuristic algorithms, such as Christofides \cite{Chen2023}, Farthest Insertion \cite{Weiler2015}, Nearest Insertion \cite{Weiler2015}, and Random Insertion \cite{Odili2015}; and learning-based solvers like Genetic Algorithm \cite{Toaza2023}, Ant Colony Optimization (ACO) \cite{Toaza2023}, Particle Swarm Optimization (PSO) \cite{Toaza2023}, and Transformer \cite{Kool2019}. The evaluation metrics is the length of the sequence of waypoints.

Lastly, for the trajectory generation module, the following algorithms were evaluated: A* \cite{Saadi2022}, AO* \cite{Saadi2022}, ILAO* \cite{Schmalz2024}, IW \cite{Junyent2021}, Best First Width Search (BFWS) \cite{Lipovetzky2017}, and Labeled Real-Time Dynamic Programming (LRTDP) \cite{Street2022}. The evaluation metrics is the length of the trajectory.

% Time
In addition, computation time was measured for all stages, with GPU measurements for segment prediction and CPU measurements for the remaining components. For this purpose, a system equipped with 2 $\times$ Nvidia GeForce RTX 4090 GPUs and a 13\textsuperscript{th} Generation Intel Core i9-13900K was used.

% Accuracy
\begin{figure}[t] % thpb
  \centering
  \includegraphics[width=\linewidth]{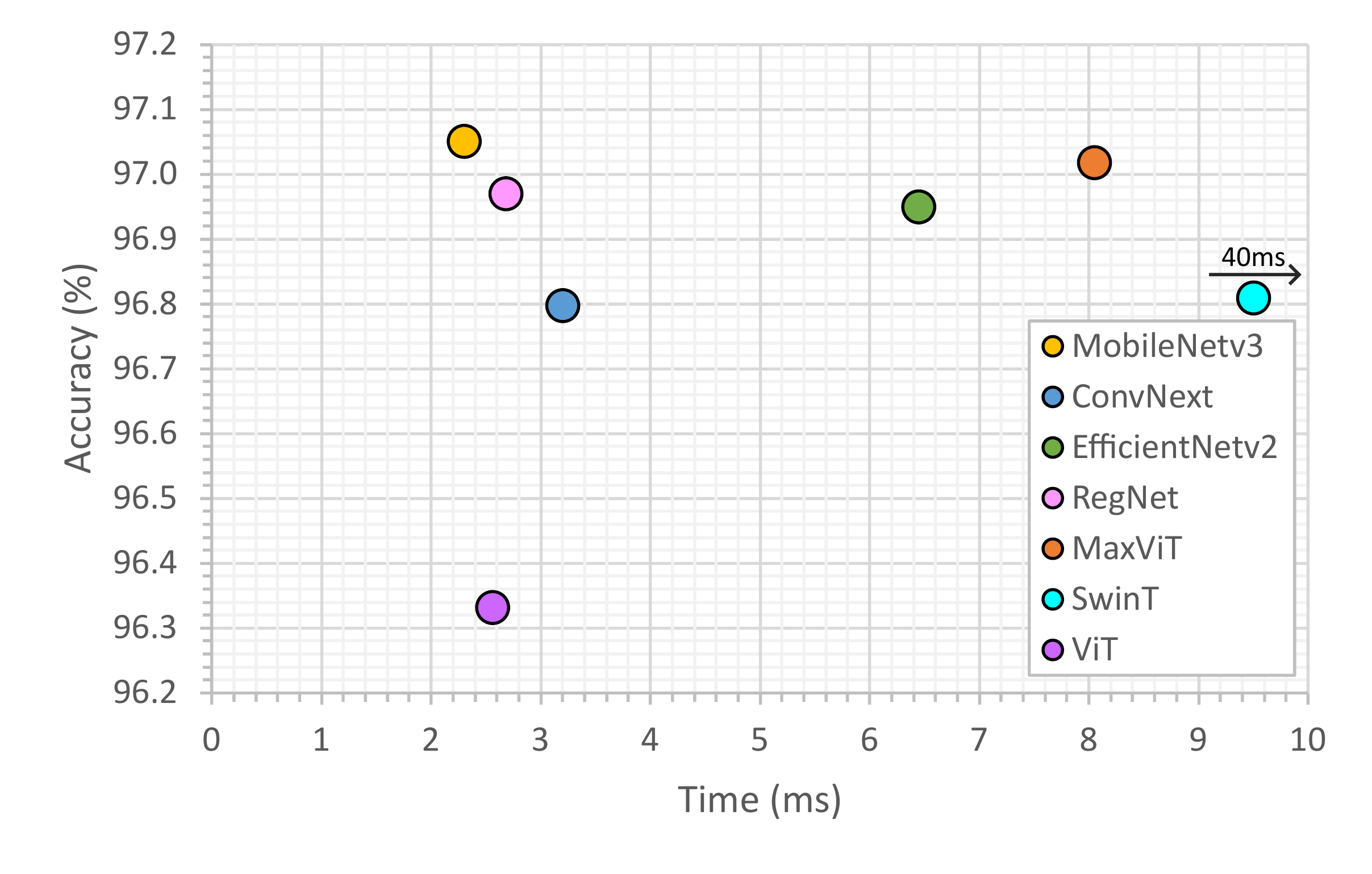}
  \caption{Performance of the segment prediction network using different backbones for feature extraction. Performance is measured in terms of accuracy and computation time (in ms).}
  \label{fig:accuracy}
\end{figure}

% Route results
\begin{table*}[t]
    \footnotesize
    \centering
    \caption{Comparison between multiple TSP solvers for waypoint sequencing. The comparison is evaluated for $n=20$, $n=50$, and $n=100$ in terms of average route length (in kilometers) and average computation time (in seconds).}
    \label{tab:route}

    % Data
    \begin{tabular}{l|cccccc}

        % Heading
        \multirow{2}{*}{Algorithms} &
        \multicolumn{2}{c}{$n=20$} &
        \multicolumn{2}{c}{$n=50$} &
        \multicolumn{2}{c}{$n=100$} \\

        % Sub-Heading
        &
        Length (km) &
        Time (s) &
        Length (km) &
        Time (s) &
        Length (km) &
        Time (s) \\
        \hline\hline

        % OR-Tools
        OR-Tools \cite{Ortools2024} &
        2172.62 $\pm$ 38.17 &
        $\phantom{0}$0.13 $\pm$ 0.01 &
        3103.86 $\pm\phantom{0}$ 62.21 &
        $\phantom{00}$2.50 $\pm$ 0.22 &
        4260.22 $\pm\phantom{0}$ 87.08 &
        10.04 $\pm$ 1.35 \\

        % Gurobi
        Gurobi \cite{Gurobi2024} &
        2175.13 $\pm$ 38.05 &
        $\phantom{0}$3.36 $\pm$ 2.10 &
        3065.83 $\pm\phantom{0}$ 59.09 &
        $\phantom{0}$51.86 $\pm$ 5.54 &
        4301.25 $\pm$ 104.31 &
        60.22 $\pm$ 0.05 \\

        % Concorde
        Concorde \cite{Concorde2006} &
        2254.82 $\pm$ 37.64 &
        \textbf{$\phantom{0}$0.01 $\pm$ 0.00} &
        3406.05 $\pm\phantom{0}$ 59.11 &
        $\phantom{00}$0.04 $\pm$ 0.00 &
        4966.44 $\pm\phantom{0}$ 81.43 &
        $\phantom{0}$0.14 $\pm$ 0.05 \\

        % B&B
        B\&B \cite{Zhang2023} &
        2183.07 $\pm$ 39.17 &
        \textbf{$\phantom{0}$0.01 $\pm$ 0.00} &
        3126.57 $\pm\phantom{0}$ 65.93 &
        $\phantom{00}$0.20 $\pm$ 0.01 &
        4234.52 $\pm\phantom{0}$ 82.93 &
        $\phantom{0}$5.79 $\pm$ 0.22 \\
         \hline

        % Christofides
        Christofides \cite{Chen2023} &
        2181.90 $\pm$ 39.04 &
        \textbf{$\phantom{0}$0.01 $\pm$ 0.00} &
        3094.74 $\pm\phantom{0}$ 57.57 &
        $\phantom{00}$0.14 $\pm$ 0.01 &
        4133.74 $\pm\phantom{0}$ 63.96 &
        $\phantom{0}$1.71 $\pm$ 0.17 \\

        % Farthest Ins.
        Farthest Ins. \cite{Weiler2015} &
        2161.94 $\pm$ 36.96 &
        $\phantom{0}$1.24 $\pm$ 0.02 &
        3043.14 $\pm\phantom{0}$ 55.47 &
        $\phantom{00}$0.85 $\pm$ 0.07 &
        4078.79 $\pm\phantom{0}$ 64.79 &
        $>$ 1000 \\

        % Nearest Ins.
        Nearest Ins. \cite{Weiler2015} &
        2160.77 $\pm$ 36.76 &
        $\phantom{0}$1.05 $\pm$ 0.02 &
        3040.62 $\pm\phantom{0}$ 53.69 &
        $\phantom{0}$88.76 $\pm$ 1.26 &
        4080.07 $\pm\phantom{0}$ 69.77 &
        $>$ 1000 \\

        % Random Ins.
        Random Ins. \cite{Odili2015} &
        2159.26 $\pm$ 36.62 &
        $\phantom{0}$0.65 $\pm$ 0.01 &
        \textbf{3028.90 $\pm\phantom{0}$ 53.33} &
        $\phantom{0}$49.35 $\pm$ 0.26 &
        \textbf{4049.15 $\pm\phantom{0}$ 66.69} &
        $>$ 1000 \\
         \hline

        % Genetic
        Genetic \cite{Toaza2023} &
        \textbf{2159.18 $\pm$ 36.63} &
        14.30 $\pm$ 0.08 &
        3031.73 $\pm\phantom{0}$ 53.32 &
        147.90 $\pm$ 1.11 &
        4059.75 $\pm\phantom{0}$ 72.39 &
        $>$ 1000 \\

        % ACO
        ACO \cite{Toaza2023} &
        2160.34 $\pm$ 36.76 &
        $\phantom{0}$0.58 $\pm$ 0.00 &
        3102.91 $\pm\phantom{0}$ 55.80 &
        $\phantom{00}$5.60 $\pm$ 0.03 &
        4275.00 $\pm\phantom{0}$ 70.23 &
        43.15 $\pm$ 0.24 \\

        % PSO
        PSO \cite{Toaza2023} &
        2303.87 $\pm$ 45.10 &
        $\phantom{0}$1.14 $\pm$ 0.01 &
        3632.83 $\pm$ 105.99 &
        $\phantom{00}$2.29 $\pm$ 0.01 &
        5107.61 $\pm$ 125.59 &
        $\phantom{0}$4.56 $\pm$ 0.04 \\

        % Transformer
        Transformer \cite{Kool2019} &
        2176.72 $\pm$ 37.18 &
        $\phantom{0}$0.02 $\pm$ 0.00 &
        3110.55 $\pm\phantom{0}$ 59.55 &
        \textbf{$\phantom{00}$0.03 $\pm$ 0.00} &
        4407.98 $\pm\phantom{0}$ 75.73 &
        \textbf{$\phantom{0}$0.07 $\pm$ 0.00} \\
        
    \end{tabular}
\end{table*}

% Trajectorires results
\begin{figure*}[t] % thpb
  \centering
    \begin{subfigure}{.41\textwidth}
        \centering
    \includegraphics[width=\linewidth]{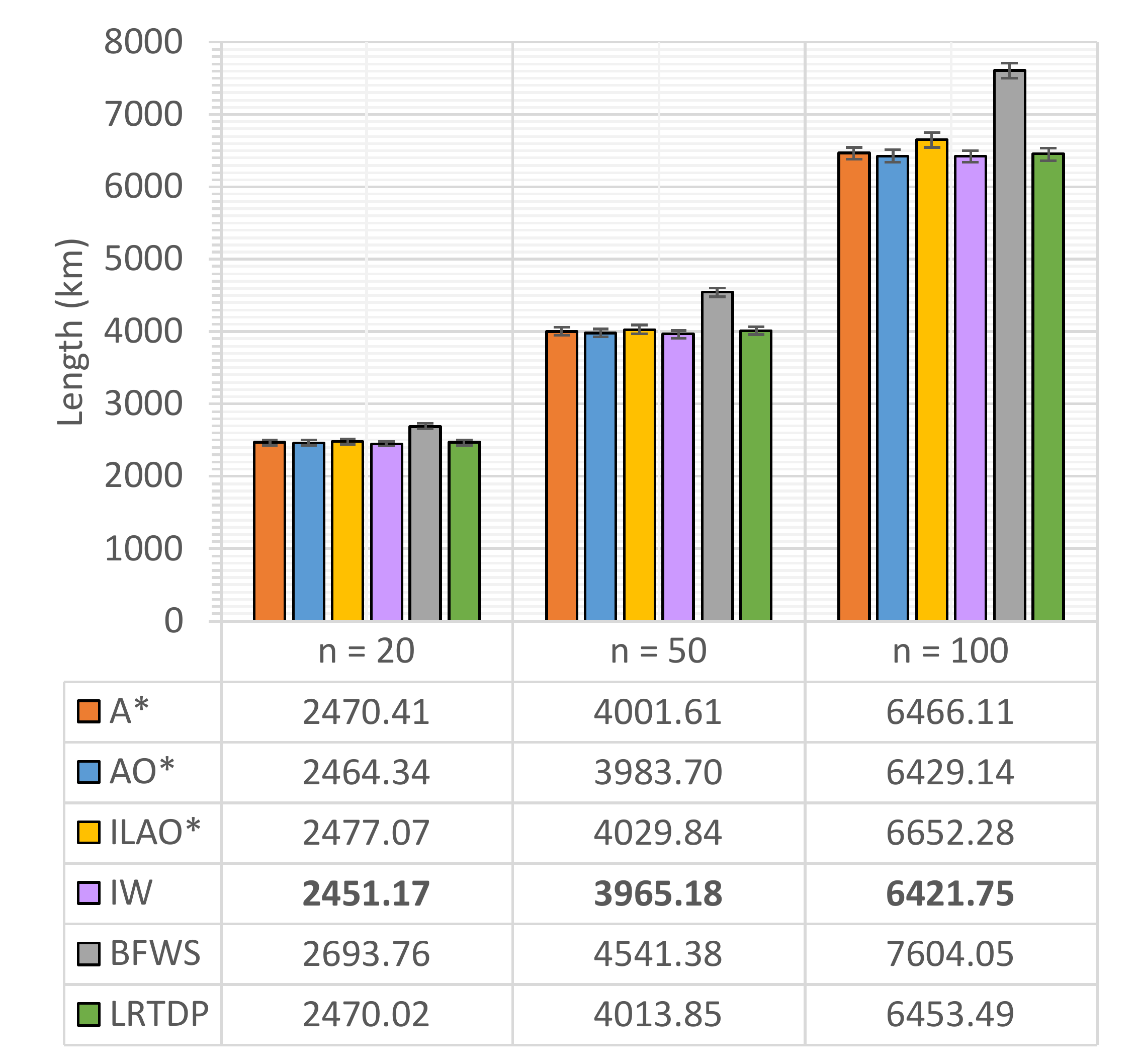}
        \caption{Path length.}
        \label{fig:path-length}
    \end{subfigure}
    \begin{subfigure}{.41\textwidth}
        \centering
        \includegraphics[width=\linewidth]{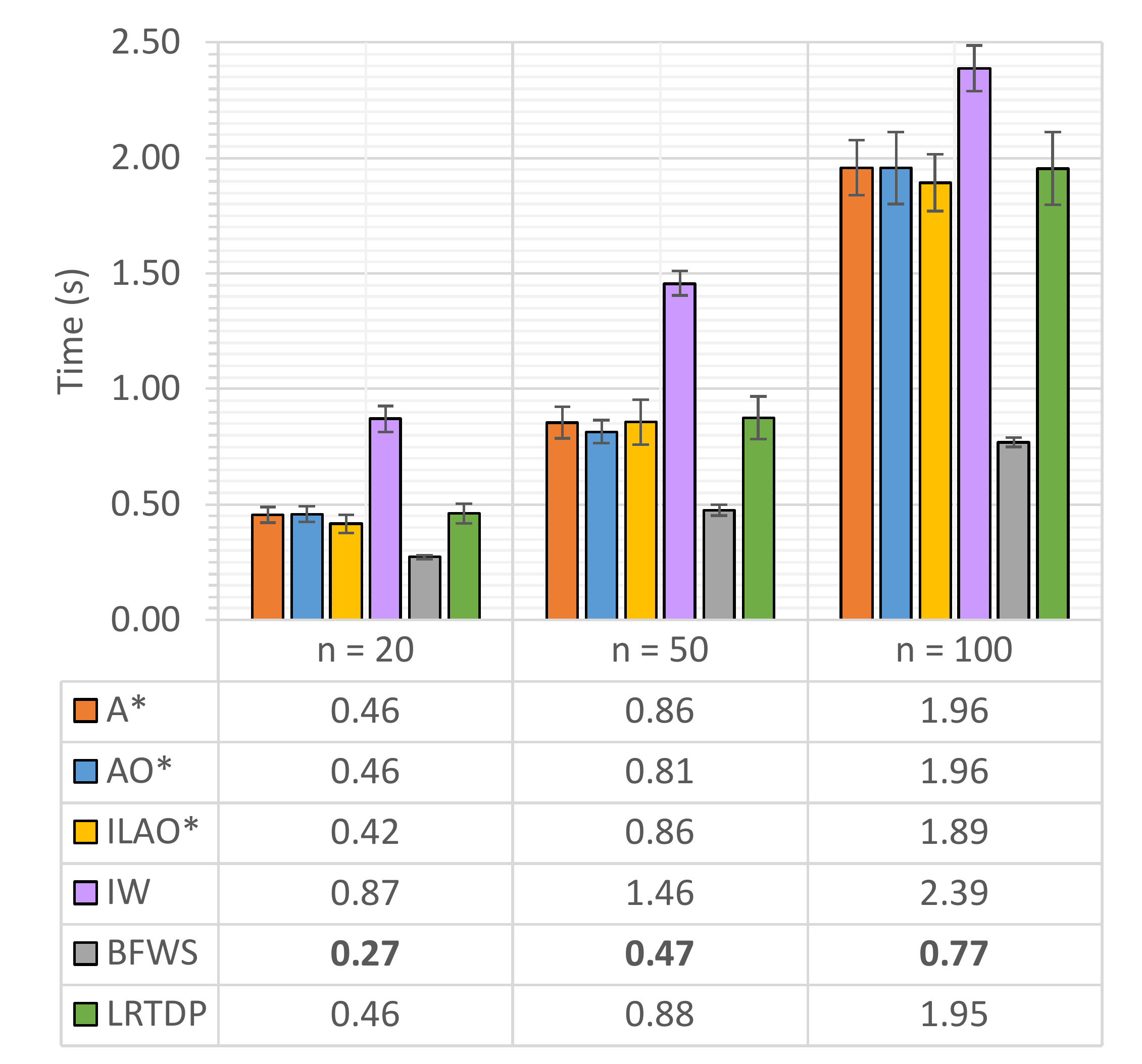}
        \caption{Computation time.}
        \label{fig:path-time}
    \end{subfigure}
  \caption{Comparison between multiple path planning algorithms for trajectory generation utilizing 3D Dubins curves. The comparison is evaluated for $n=20$, $n=50$, and $n=100$ in terms of (a) average path length (in kilometers) and (b) average computation time (in seconds). Both subfigures include 95\% confidence intervals for each algorithm.}
  \label{fig:path}
\end{figure*}

% Segment prediction
The results of the segment prediction stage (see Fig. \ref{fig:accuracy}) reveal no significant differences in accuracy between the backbone architectures tested. All models exhibit comparable performance, with accuracy values clustering within a narrow range. The 95\% confidence intervals for accuracy remain around $\pm0.6$, while for computation time they do not exceed $\pm1$ ms. These results indicate a clear saturation in performance, suggesting that the proposed approach is robust to the choice of backbone network, and also that increasing complexity of the backbone does not yield substantial benefits in accuracy.

% Waypoint sequencing
The results of the waypoint sequencing stage, presented in Tab. \ref{tab:route}, highlight substantial differences in performance between the algorithms tested. The most accurate methods, in terms of minimizing route length, are the Genetic Algorithm (GA) and the insertion heuristics (IH), either Farthest, Nearest, or Random Insertion variants. However, the GA requires long processing time to generate solutions. The insertion methods, while efficient for small-scale problems ($n=20$), suffer from poor scalability, with computation times increasing drastically as the problem size grows. On the other hand, the fastest algorithms are Transformer and Concorde, which can provide solutions in less than 0.15 seconds for the largest scenarios ($n=100$), unlike GA and IH ($>$1000 seconds). However, the Transformer-based approach outperforms Concorde in terms of route length, demonstrating its effectiveness in producing the best trade-off between average-quality solutions and computational cost.

% Trajectory generation
Finally, the evaluation of trajectory generation algorithms, illustrated in Fig. \ref{fig:path}, provides further insights into the overall system performance. Fig. \ref{fig:path-length} shows the total path length obtained after connecting all segments predicted by the segment prediction network following the sequence determined by the Transformer-based routing network. And Fig. \ref{fig:path-time} presents the computation times for each trajectory generation method. The results indicate minimal differences in path length across most algorithms, with the exception of BFWS, which yields longer trajectories but is computationally faster. The IW algorithm, while achieving comparable path lengths to the other methods, exhibits significantly higher computation times, suggesting an inefficiency in its search process. Among the remaining algorithms, no substantial differences are observed, suggesting that trajectory generation performance is relatively consistent across different approaches.

% Limitations
Finally, some perceived limitations of the system include the assumption of a fixed set of static targets and offline operation, with no mechanisms for dynamic replanning or adaptation to changing mission conditions. Moreover, the evaluation was limited to natural terrain elevations, excluding other environments such as urban areas. Additionally, the system’s stages (waypoint sequencing, segment prediction, and trajectory generation) operate independently, which may lead to suboptimalities due to the lack of joint optimization.

\section{Conclusions}
\label{sec:conclusions}

This work addresses the challenge of multi-target path planning for SAR-equipped aircraft by introducing a holistic system that combines three stages: a Transformer-based waypoint sequencer to enable rapid and scalable route planning; a novel straight-flight segment prediction network based on a MobileNetv3 backbone to ensure efficient prediction of terrain-aware flight segments, prioritizing visibility at minimal altitudes; and a trajectory generation module that uses A* search with 3D Dubins curves to guarantee feasible and smooth paths. Contrary to other state-of-the-art works that mainly focus on optimizing trajectories for individual targets, the proposed system provides a comprehensive solution to the entire multi-target problem. Evaluations reveal that the proposed system is capable of predicting accurate paths for robust SAR-imaging, and that it is suitable for real-world applications, particularly in scenarios requiring rapid decision-making. Future work should focus on obstacle avoidance, dynamic path replanning, and joint optimization across all stages in a unified framework.

% Acknowledgement
\section*{ACKNOWLEDGMENT}
This work was supported in part by the Comunidad de Madrid under project TEC-2024/COM-322 (IDEALCV-CM), in part by MCIU/AEI/10.13039/501100011033 of the Spanish Government under project PID2023-148922OA-I00 (EEVOCATIONS), and in part by MCIN/AEI/10.13039/501100011033 and the “European Union NextGenerationEU/PRTR” under grant TED2021-130225A-I00 (ANEMONA).

% References
\bibliographystyle{IEEEtran}
\bibliography{refs}

\end{document}